\pdfoutput=1
\documentclass[11pt]{article}
\usepackage[final]{acl}
\usepackage{times}
\usepackage{latexsym}
\usepackage{booktabs}
\usepackage[T1]{fontenc}
\usepackage[utf8]{inputenc}
\usepackage{microtype}
\usepackage{inconsolata}
\usepackage{graphicx}
\usepackage{subcaption}
\usepackage{multicol}
\usepackage{multirow}
\usepackage{tcolorbox}
\usepackage{amsmath}
\usepackage{algorithm,algorithmic}
\usepackage{pbox}
\usepackage{enumitem}
\definecolor{backcolor}{RGB}{232, 242, 255}
\usepackage{pifont}
\newcommand{\cmark}{\ding{51}}
\newcommand{\xmark}{\ding{55}}
\usepackage[table]{xcolor}

\title{DivLogicEval: A Framework for Benchmarking Logical Reasoning \\ Evaluation in Large Language Models}

\author{
    Tsz Ting Chung\textsuperscript{1} \quad
    Lemao Liu\textsuperscript{2} \quad
    Mo Yu\textsuperscript{3} \quad
    Dit-Yan Yeung \textsuperscript{1} \\
    \textsuperscript{1}The Hong Kong University of Science and Technology \\
    \textsuperscript{2}Fudan University \quad
    \textsuperscript{3}WeChat AI, Tencent \\
    \texttt{ttchungac@connect.ust.hk} \quad
    \texttt{lemaoliu@gmail.com} \\
    \texttt{moyumyu@global.tencent.com} \quad
    \texttt{dyyeung@cse.ust.hk} \\
}

\begin{document}
\maketitle
\begin{abstract}
Logic reasoning in natural language has been recognized as an important measure of human intelligence for Large Language Models (LLMs). 
Popular benchmarks may entangle multiple reasoning skills and thus provide unfaithful evaluations on the {\em logic} reasoning skill. Meanwhile, existing logic reasoning benchmarks are limited in language diversity and their distributions are deviated from the distribution of an ideal logic reasoning benchmark, which may lead to biased evaluation results. This paper thereby proposes a new classical logic benchmark DivLogicEval, consisting of natural sentences composed of diverse statements in a counterintuitive way.
To ensure a more reliable evaluation, we also introduce a new evaluation metric that mitigates the influence of bias and randomness inherent in LLMs. Through experiments, we demonstrate the extent to which logical reasoning is required to answer the questions in DivLogicEval and compare the performance of different LLMs in conducting logical reasoning. 
\end{abstract}

\section{Introduction}
\label{sc:intro}
Large Language Models (LLMs) have been evaluated across diverse reasoning capabilities, including narrative reasoning~\cite{yu2025preludebenchmarkdesignedrequire, zhou-etal-2025-essence}, mathematical reasoning~\cite{hendrycksmath2021, cobbe2021training}, inductive reasoning~\cite{yu-etal-2025-stochastic, chollet2019measureintelligence},  and logical reasoning~\cite{PrOntoQAOOD, Yu2020ReClor}. Among these, logical reasoning has long been regarded as a key indicator of human intelligence, frequently employed in contexts such as academic admissions, employment screening, and civil service recruitment. Many globally recognized standardized tests, such as the LSAT, GMAT, civil service examinations, and general aptitude tests, include a substantial number of logic-based questions.

\begin{table}[ht!]
{\small
\centering
\begin{tabular}{@{}l|ccc@{}}
\toprule
                & Diversity & Logic-centered  \\ \midrule

ReClor  & \cmark    &   \xmark  \\ 
LogiQA2  & \cmark    &   \xmark  \\ \midrule
RuleTaker & \xmark     &   \cmark   \\
LogicNLI & \xmark   &   \cmark   \\
FOLIO & \xmark     &   \cmark   \\
RobustLR & \xmark    &   \cmark   \\ 
PrOntoQA-OOD & \xmark  &   \cmark   \\ \midrule
\rowcolor{backcolor} 
DivLogicEval   & \cmark    &  \cmark    \\ \bottomrule
\end{tabular}
\caption{Comparison between the proposed DivLogicEval and existing logic reasoning benchmarks. DivLogicEval demonstrates great language diversity and provides a logic-centered evaluation that isolates the impact of logical reasoning from factors like commonsense reasoning and pretraining shortcuts.}
\label{tb: intro}
}
\end{table}

\begin{figure*}
    \centering
    \includegraphics[width=\linewidth]{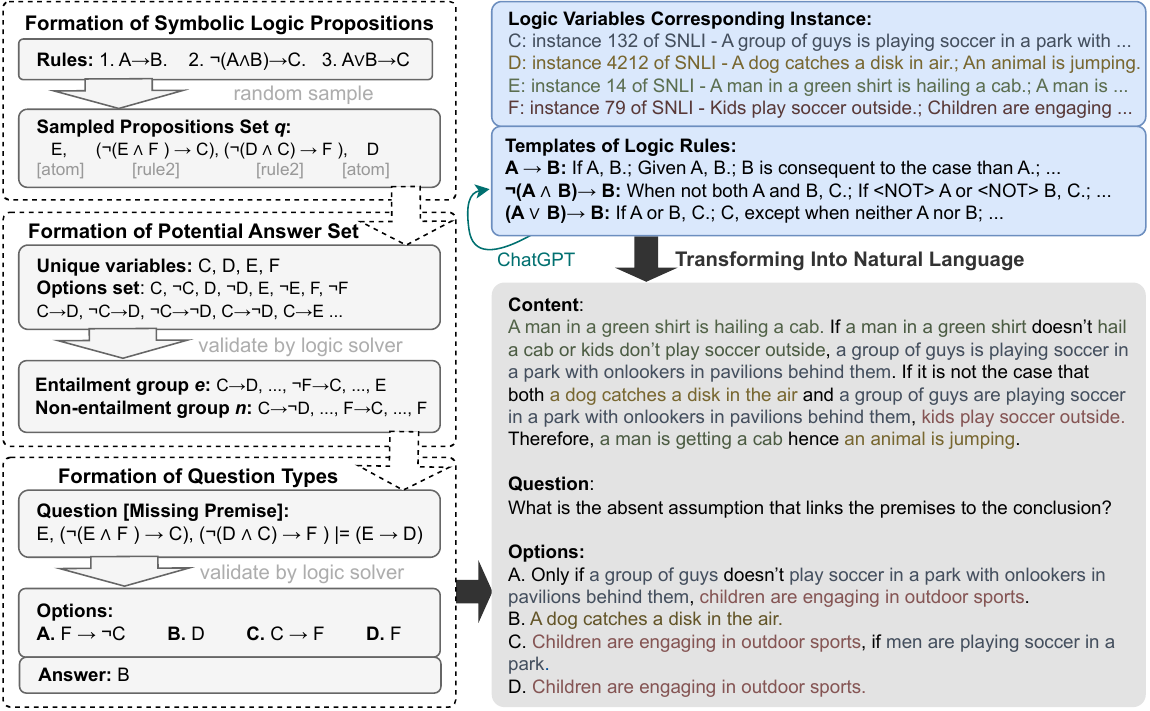}
  \caption{
    Illustration with the construction process.}
 \label{fig:3ques}
 \vspace{-5mm}
\end{figure*}

Although some popular benchmarks such as ReClor~\cite{Yu2020ReClor} and LogiQA~\cite{Liu2020LogiQAAC} exist for evaluating logical reasoning~\cite{Yu2020ReClor,Liu2020LogiQAAC,10174688,huang2023discourse}, they do not solely concentrate on logic reasoning. In fact, various reasoning (e.g., commonsense reasoning and logic reasoning) skills may entangle in these benchmarks and some other reasoning skills can make non-negligible contributions to solving the task. For example, as shown in experiments (see \S3.1), by simply reducing the model's ability to utilize logical reasoning through prompting, LLM surprisingly yields better performance on ReClor and LogiQA. Consequently, such benchmarks may overestimate the logic reasoning abilities of LLMs.

Meanwhile, some efforts have been made to create datasets based on classical logic~\cite{hahn2021teaching,pi-etal-2022-reasoning,sanyal-etal-2022-robustlr}. For example, FOLIO is equipped with human annotations of classical logic~\cite{han2022folio} but it only contains about two hundred instances due to annotation cost. Other datasets are synthetically generated by predefined templates over classical logic expressions~\cite{ruletaker,tian-etal-2021-logicnli,PrOntoQA}. Unfortunately, the language variety within these benchmarks is limited due to the high expenses associated with manually annotating diverse sentences, while their predefined templates are unable to ensure sentence diversity. Therefore, they inherently suffer from an important limitation in distribution bias. As quantified in Section 3.2, for the most diverse synthetic dataset, its vocabulary size is only about two hundred, and their data distribution is largely deviated from the distribution of natural language. As a result, such test datasets may induce a biased evaluation result according to the statistic sampling principle~\cite{cochran1977sampling,lohr2021sampling}.

In this paper, we thereby propose a \textbf{Div}erse \textbf{Logic}al Reasoning \textbf{Eval}uation (DivLogicEval) benchmark, 
which includes a {\em diverse} evaluation dataset with a rich vocabulary. 
Specifically, as illustrated in Figure~\ref{fig:3ques}, the key idea to constructing DivLogicEval includes two steps (\S 2.1):
we first sample a classical logic expression, which is verifiable by an external logic solver from a predefined set of symbolic logic propositions; and then the logic expression is transformed into natural language by instantiating its variables with {\em diverse} natural sentences and concatenating them with counterintuitive sentence connectives.

Moreover, we propose a new evaluation metric for DivLogicEval which posses additional benefits over the existing evaluation metrics (\S 2.2). 
Our further analysis demonstrates the effectiveness of DivLogicEval: DivLogicEval not only concentrates more on the logic reasoning skill than ReClor and LogiQA but also surpasses existing \textit{Logic}-aware benchmarks in language diversity (\S 3), as illustrated in Table \ref{tb: intro}. 
 
Finally, we evaluate popular LLMs include open-sourced and closed-sourced LLMs on DivLogicEval with some interesting findings (\S 4). 

\section{DivLogicEval Benchmark}

\subsection{Dataset Construction}

\paragraph{Overview.}
DivLogicEval is a multiple-choice MRC dataset for ease and effectiveness in evaluation, and it consists of three components: content, passage, and (four) options, with only one of them being correct. DivLogicEval is a synthetic dataset constructed based on verifiable propositional logic in a counterintuitive manner. 
The construction method comprises several steps as illustrated in Figure~\ref{fig:3ques}. We first sample a propositional logic expression that can be verified by an external logic solver, thereby creating a predefined set of symbolic logic propositions. Subsequently, the logic expression is transformed into natural language by instantiating its variables with diverse natural sentences. Finally, these sentences are concatenated with connectives in a counterintuitive manner.
The detailed construction process is described in the rest of this subsection. 

\paragraph{Formation of Symbolic Logical Propositions.}
\label{sc: proposition}
Two to three logic atom units are sampled from a set of eight possible variables (i.e., `A', `B', ..., `H') iteratively while the probability of sampling the variables will drop in case it is being selected. The selected variables are then incorporated into the three implication rules that are commonly used in previous research to address \cite{wang-etal-2022-logic,li-etal-2022-adalogn,locsgn} or generate \cite{ruletaker,sanyal-etal-2022-robustlr} logical reasoning benchmarks. 
The sampled propositions are concatenated to form the content of the MCQA.
\begin{subequations}
\begin{align}
((A \rightarrow B) &\rightarrow (\neg B \rightarrow \neg A)) \\
((\neg (A \land B) \rightarrow C) &\rightarrow (\neg A \rightarrow C)) \\
(((A \lor B) \rightarrow C) &\rightarrow (A \rightarrow C))
\label{eq:rules}
\end{align}
\end{subequations}

\paragraph{Formation of Potential Answer Set.} The potential answer set is constructed with all possible pairs of variables with inference relation as well as the single atom variable, including the consideration of the negated variables. With an external logic validator, the options set can be divided into the entailment group $e$ and the non-entailment group $n$. To ensure at least two propositions in the content are necessary to derive the correct answer, the variable pairs in the entailment group $e$ that can be directly derived from a single proposition $q_i$ are filtered out.

\paragraph{Formation of Question Types.}
DivLogicEval is designed with three similar question types to the renowned GMAT examination in this domain.
The content part for different question types is normally as the premise, except for ``Missing Premise''.

\vspace{2mm}
\hspace{-4.8mm} \textbf{\textit{3c1e:}} The content fails to imply three of the options while implying the remaining one.

\vspace{1mm}
\hspace{-4.7mm} \textbf{\textit{3e1c:}} The content implies three options while failing to imply the remaining one. 
\vspace{1mm}

\hspace{-4.9mm} \textbf{\textit{Missing Premise:}} 
 The content part is modified to combine with a valid conclusion from the entailment group. The necessary proposition in the premise, which guarantees the premise leads to the conclusion, is then removed. The removed proposition then becomes the correct option. Meanwhile, the remaining three options are drawn from the non-entailment group, with further validation conducted by an external validator. A sample instance of this question type is presented in Figure \ref{fig:3ques}.

\paragraph{Transforming Into Natural Language.}
To ensure the richness of language diversity, each logical variable is replaced by simple sentences sourced from SNLI \cite{2018-snli} and MNLI datasets \cite{williams-etal-2018-broad-mnli}. Inappropriate sentences from these datasets are filtered by predefined rules to ensure the language quality and details are provided in the supplementary materials. 

To generate natural language templates for the three inference rules, GPT-3.5 is employed with several sample templates provided as input. For instance, one of the templates for expressing logical implication, "A implies B," can be used. 

The finalized text is subsequently subjected to a grammar check by passing it through GPT-3.5 once again. The ratio calculated by dividing the length of the longest common substring between the original text fragment and the modified fragment over the maximum fragment length is applied. To prevent excessive modifications made by GPT-3.5, only proposed changes with a ratio greater than 0.5 are retained. Multiple trials are conducted, and the result with the smallest amount of modification is adopted. To further maintain the quality of the testing set, we manually review and approve the changes suggested by GPT-3.5.

\paragraph{Remarks on Language Diversity.}
As presented in section \ref{sc:intro}, language diversity is an important factor in making evaluation on our benchmark unbiased and reliable. To make our benchmark diverse, we take language diversity into account in the following four aspects when transforming logic expressions into natural language. 

(1) We instantiate each variable in logic expressions with a natural sentence. Since a natural sentence itself is diverse enough, the diversity of our benchmark can therefore be achieved. (2) Since an expression may contain the same variable multiple times, we instantiate such a variable with different sentences with NLI datasets where multiple expressions with the same or contradicting meanings exist. (3) GPT-3.5 is utilized to ensure the templates for combining different natural language expressions are diversified. (4) Sentence negation is performed to add diversity to an existing natural language expression. 

\subsection{Evaluation Metric}
\paragraph{Issues of Existing Metrics.}
On the existing benchmarks such as ReClor, two popular metrics (i.e., Accuracy and Circular) are used for evaluation. 
`Accuracy' measures the accuracy of the original instance. Instead, `Circular' evaluates all mutants of an instance by shifting the order of the options in a circular way~\cite{mmbench}, and an instance is considered correct only if all mutants with different options in different positions are answered correctly.~\footnote{For example, if the four options are denoted $o_1$, $o_2$, $o_3$, and $o_4$, and the original instance is denoted by $(o_1, o_2, o_3, o_4)$, then the four created mutants are $(o_1, o_2, o_3, o_4)$, $(o_2, o_3, o_4, o_1)$, $(o_3, o_4, o_1, o_2)$, and $(o_4, o_1, o_2, o_3)$. An instance is considered correct only if all the mutants are answered correctly.} 

Unfortunately, we find that both existing metrics lack faithfulness. Specifically, when comparing GPT-4 and Gemini in a zero-shot setting on DivLogicEval, Gemini-1.0-pro may outperform GPT-4-turbo in a single trial in terms of accuracy, as shown in Table~\ref{tb:acc}. However, this trend does not hold across other trials, as indicated in Table~\ref{tb:cv}. In a similar experimental setting focused on logical reasoning with distracting rules, Chen~\shortcite{chen2024premiseordermattersreasoning} demonstrates that GPT-4 consistently outperforms Gemini-1.0-pro. Using the same LLMs, we obtain the same conclusion with Circular and PartialCircular metrics. Furthermore, as shown in Table~\ref{tb:cv}, the coefficient of variance of PartialCircular is significantly better than that of Circular calculated over five independent runs. This reduces the likelihood of drawing inconsistent conclusions across runs and demonstrates the advantage of introducing PartialCircular. 

\begin{table}[htbp]
\centering
\small
\begin{tabular}{@{}l|c|c@{}}
\toprule
        & Gemini (run3) & GPT-4 \\ \midrule
Accuracy (ACC)       & 32.4    & 32.2    \\
Circular (CIR)        & 8.0     &  12.3    \\
PartialCircular (PC)  &  18.1    & 22.1  \\\midrule
\end{tabular}
\caption{Accuracy (ACC) is not consistent with Circular (CIR) and PartialCircular (PC). Gemini-Pro is comparable to GPT-4 in terms of ACC but it is not in terms of CIR and PC.}
\label{tb:acc}
\end{table}

\begin{table}[htbp]
\centering
\small
\begin{tabular}{@{}l|ccccc|c@{}}
\toprule
                & run1 & run2 & run3 & run4 & run5 & CV\\ \midrule
ACC       & 30.0    & 32.0    &  32.4   & 30.1    &  32.0  & 3.3 \\
CIR        & 7.4     &  8.1    &  8.0    & 8.0     &  9.0   & 6.3 \\
PC  & 17.6    & 18.8    &  18.1   & 18.5    & 19.1   & 3.1 \\\midrule
\end{tabular}
\caption{Coefficient of variance (CV) for Circular (CIR) is much larger than those for Accuracy (ACC) and PartialCircular (PC). CV is calculated on five runs of Gemini-Pro.}
\label{tb:cv}
\end{table}

\paragraph{PartialCircular.}
The intuition behind Circular is to evaluate the model’s confidence alongside correctness, rather than focusing solely on accuracy. With a higher probability in a particular option, it reflects the model's confidence beyond random guessing. In cases where the model randomly guesses the answer (i.e., equal probabilities among all options), even a correct guess in one instance should not earn credit, as this would overestimate the model’s true understanding or reasoning ability.

Inspired by Circular, we propose an alternative metric PartialCircular. Similar to Circular, PartialCircular also takes into account all mutants for an instance. However, PartialCircular would assign a non-zero value to an instance even if there is an incorrect prediction for some of its mutants. Formally, the computation of PartialCircular per instance is designed as:

\begin{equation}
    \frac{c}{4} \cdot (1+\sum_{o} p(o)\log_4 p(o))
    \label{eq:pc}
\end{equation}
where $c$ is the number of mutants being answered correctly, $p(\cdot)$ is the frequency distribution of the predicted options for all the mutants.
For example, suppose there are four mutants$(o_1, o_2, o_3, o_4)$, $(o_2, o_3, o_4, o_1)$, $(o_3, o_4, o_1, o_2)$, and $(o_4, o_1, o_2, o_3)$, and their predicted options are $o_1$, $o_1$, $o_3$ and $o_4$, respectively. If the ground-truth option is $o_1$ for all four mutants, then $c=2$, $p(o_1)=2/4$, $p(o_2)=0$, $p(o_3)=P(o_4)=1/4$ and thus PartialCircular value is $0.125$. The intuition behind the entropy in Eq.~\eqref{eq:pc} is to penalize predicted frequency distributions with high entropy, as higher entropy indicates greater uncertainty. Additionally, we provide an alternative equation for scenarios where partial credit can be awarded for a correct answer, even under the case of random guessing. For further details, see Appendix \ref{sec:alteq}.

\begin{table}[ht]
\centering
\small
{
\begin{tabular}{@{}l|l|cc|c@{}}
\toprule
             & & Origin & NoLR & $\Delta$ \\ \midrule
\multirow{3}{*}{DivLogicEval} & ACC          & 32.2 & 28.6 & -3.6\\
& CIR          & 16.4 & 15.8 & -0.6 \\
& PC                 & 6.3   & 5.2 & -1.1
 \\\midrule
\multirow{3}{*}{ReClor} & ACC         & 57.3 & 59.8 & +2.5\\
& CIR          & 32.1  & 35.0 & +2.9\\
& PC             & 45.1  & 48.0 & +2.9
 \\\midrule
\multirow{3}{*}{LogiQA2} & ACC         & 51.9 & 53.2 & +1.3\\
& CIR          & 15.9  & 20.7 & +4.8\\
& PC             & 28.0  & 33.6 & +5.6
 \\\bottomrule
\end{tabular}
\caption{Performance comparison between prompting GPT-3.5 to answer without using logical reasoning (NoLR) and using the original prompt (Origin). $\Delta$ refers to the difference between Origin and NoLR.}
\label{tb:singlereasoning}
}
\end{table}

\begin{table*}[ht]
{\small
\centering
\begin{tabular}{@{}lc|cc|ccccc@{}}
\toprule
                       & {\footnotesize DivLogicEval}  & {\footnotesize ReClor} & {\footnotesize LogiQA{\scriptsize 2}} & {\footnotesize RuleTaker} & {\footnotesize LogicNLI}  & {\footnotesize FOLIO}                                                                             & {\footnotesize RobustLR} & {\footnotesize  PrOntoQA{\scriptsize -OOD}}  \\ \midrule
\# of instances       & 900           & 1000   & 1470    & 100k      & 2000      & 227                                                                               & 120k  & 1450  \\
vocabulary size        & 6748          & 7785   & 13546   & 67        & 241       & 1140                                                                              & 47    & 108  \\ 
KL & \textbf{1.87} & \textbf{1.44} & \textbf{1.32} & 4.29 & 4.62 & 2.77 & 6.28 & 6.24 \\
\bottomrule
\end{tabular}
\caption{\textbf{Analysis of distribution bias among different datasets (i.e., testing set).} KL divergence between the vocabulary frequency distributions in the testing set and wiki is presented as the metric of distribution bias.}
\label{tb:overviewd}
}

\end{table*}
\section{Effectiveness of DivLogicEval}
 DivLogicEval effectively decreases the dependency on other reasoning abilities and mitigates the risk of distribution bias by providing a more diverse dataset. 

\subsection{Reliance on Logic Reasoning}
DivLogicEval is proposed on top of logic reasoning in a counterintuitive nature and we hope that it has the potential to prevent the task from being solved with other reasoning abilities rather than logic reasoning ability. Theoretically, it is intractable to disentangle the logic reasoning ability from all types of reasoning abilities. In practice, we design a simple experiment to roughly verify our hypothesis by using the prompting technique in LLMs. Specifically, we prompt an LLM to disable its logic reasoning ability during the inference, i.e., we add  `Please try your best to answer correctly without performing any logical reasoning' (NoLR).

We conduct experiments on GPT-3.5 and compare DivLogicEval with two popular benchmarks ReClor and LogiQA2. The results are shown in Table \ref{tb:singlereasoning}. From Table \ref{tb:singlereasoning}, we can see that the NoLR system is worse than the original system on DivLogicEval. However, the decrease is not as large as one expected. One possible reason is that the NoLR prompt can not fully disable the logic reasoning ability for GPT-3.5 but decreases its usage of logic reasoning to some extent. In contrast, to our surprise, the NoLR system is even better than the original system on both ReClor and LogiQA2. This supports our hypothesis that DivLogicEval relies more heavily on logical reasoning compared to ReClor and LogiQA2. At the same time, it highlights that other forms of reasoning, along with the pretrained knowledge of LLMs, make a non-negligible contribution to the performance of existing linguistically diverse benchmarks ReClor and LogiQA2.

\vspace{1mm}
\subsection{Distribution Bias}

According to the statistical sampling principle, testing instances should be randomly sampled in an independent identically distributed manner. Otherwise, the evaluation result on such test data will be biased and unreliable~\cite{cochran1977sampling,lohr2021sampling}.
Unfortunately, there are some inevitable biases in the existing logic reasoning benchmarks as well as our proposed benchmark during their construction~\cite{han2022folio,ruletaker,tian-etal-2021-logicnli,PrOntoQA}. For example, FOLIO dataset~\cite{han2022folio} is from a particular domain and may be biased to the preferences of annotators for manual modifications;  Ruletaker, LogicNLI and PrOntoQA are generated by a set of predefined (biased) templates~\cite{ruletaker,tian-etal-2021-logicnli,PrOntoQA}. Consequently, it is crucial to study the effects of distribution bias on the classical logic benchmarks.

Since the exact distribution for the data is unknown, it is intractable to exactly measure the distribution bias for a classical logic dataset. 
Instead, we have some relaxed requirements for an ideal classical logic dataset. 
For example, as the essence of natural language sentences, such data should be diverse and its distribution should be similar to the distribution of general data such as wikipedia. Based on this requirement, we conduct an experiment to approximately measure the distribution bias. 
We first randomly select a subset noted by wiki-subset from wiki dataset~\footnote{Subset ``20220301.simple'' is used}. Then we calculate the KL divergence between the frequency distribution of the wiki subset and that of each test set from several classical logic benchmarks. To ensure a fair comparison, we control the number of tokens in the wiki subset to be similar to that of the smallest test set (i.e., FOLIO) among those classical logic benchmarks.  

 As a reference, we also calculate the KL divergence for non-classical logic benchmarks (Reclor and LogiQA2). The result presented in Table \ref{tb:overviewd} shows: 
 \begin{enumerate}
    \setlength\itemsep{0em}
    \item Our benchmark is significantly more diverse than existing classical logic benchmarks in vocabulary.
    \item Our benchmark achieves the lowest KL divergence compared to the classical logic benchmarks.
\end{enumerate}

\subsection{Effect of Contamination}
\label{sc:contam}

The issue of data contamination has been increasingly recognized in the research community due to its detrimental impact on model evaluation. LLMs can solve existing benchmarks through memorization of pretrained data rather than reasoning, leading to an overestimation of their performance by the current evaluator. 
The counterintuitive nature of our benchmark offers additional advantages in potentially preventing data contamination. We design two experiments to analyze the effect of DivLogicEval regarding to data contamination. 

First, we directly examine the contamination level of DivLogicEval with regard to the current GPT-3.5 following Deng~\shortcite{deng2024investigating} approach. The contamination level is measured by the exact match rate between the predicted option and its original option which is incorrect. We find that the exact match rate is only 0.2\%, which demonstrates that DivLogicEval remains uncontaminated in GPT-3.5, despite our dataset being sourced from popular NLI benchmarks.

Second, we study the effect of contamination when both SNLI and MNLI datasets are covered in the pretraining dataset of open-source LLMs. This situation is likely to occur as the LLMs continue to expand their pre-training datasets. To this end, an open-sourced LLaMA2 is used for additional pre-training on both SNLI and MNLI datasets using LLaMA-Factory~\cite{zheng2024llamafactory} with LoRA tuning. We compare the performance gap between the additionally pretrained model and the original model on DivLogicEval. This performance gap serves as an indicator of the effect of contamination. Our results presented in Table~\ref{tb:contam} demonstrate that the impact of contamination on DivLogicEval is negligible compared to the effect on SNLI.

\begin{table}[htbp]
\centering
{\small
\begin{tabular}{@{}l|ccc|c@{}}
\toprule
           & origin & tuned(1x) & tuned(10x) & $\Delta$ \\\midrule
SNLI       & 41.1   & 53.8  & 53.8 & +12.7 \\\midrule
DivLogicEval & 26.9   & 28.1  & 28.1 & +1.2
 \\\bottomrule
\end{tabular}
\caption{Task performance comparison between SNLI and DivLogicEval in llama-2-7b (origin) and llama-2-7b with further pretraining on SNLI (tuned) for one (1x) and ten epoch (10x). $\Delta$ refers to the difference between tuned and origin.}
\label{tb:contam}
}
\vspace{-2mm}
\end{table}

\section{Evaluating LLMs on DivLogicEval}
\subsection{Settings}
\paragraph{Datasets}
DivLogicEval is a multiple-choice dataset, comprising four options, with one option being the correct answer. The dataset comprises a total of 12,589 instances, distributed as follows: 4196 instances corresponding to '3c1e', 4195 instances corresponding to '3e1c', and 4198 instances corresponding to  'missing premise'.

The dataset is partitioned into unpolished and polished sets (i.e., testing set). The human post-editted polished set consist of 900 instances, with the class balance being maintained within each set. The distinct vocabulary size of DivLogicEval, determined using the nltk tokenizer, is comparable to that of complex datasets such as ReClor and LogiQA2 as illustrated in Table \ref{tb:overviewd}. Additionally, it is significantly larger than that of classical logic reasoning datasets. 

\vspace{-1.5mm}

\paragraph{Configurations.}
As the scale of model size increases, LLMs inherently gain the capability to handle various natural language tasks in a zero-shot setting. \cite{wei2022emergent,kojima2022large}. In addition to studying their performance under a zero-shot setting, we further investigate their performance in a few-shot setting~\cite{fewshot, wei-etal-2023-symbol, chung-etal-2024-selection} by providing examples for guidance. Under the three-shot setting, the models are provided with three examples to facilitate their learning process prior to answering each question. An example is ensured for each question type, and they are sampled from the unpolished set. The models studied include Mixtral ({\small \texttt{mixtral-8x7B-instruct-v0.1}}) \cite{jiang2023mistral}, LLaMA 3.3 ({\small \texttt{llama-3.3-70b-instruct}}) \cite{touvron2023LLaMA}, Qwen 2.5 ({\small \texttt{qwen-2.5-72b-instruct}}) \cite{qwen2025qwen25technicalreport}, Gemini ({\small \texttt{gemini-1.5-pro}}) \cite{geminiteam2023gemini}, GPT-3.5 ({\small \texttt{gpt-3.5-turbo}}), GPT-4 ({\small \texttt{gpt-4-1106-preview}}), GPT-4o ({\small \texttt{gpt-4o-2024-05-13}}) and o1-preview ({\small \texttt{o1-preview-2024-09-12}}) \cite{openai2023gpt4}  They are prompted with the instruction ``You need to answer in the form of Answer: \(<\)A/B/C/D\(>\)''. We evaluate all these systems in terms of all three metrics mentioned in section 2.1, i.e., Accuracy, Circular, and the proposed PartialCirular.

\begin{figure*}[ht]
    \centering
    \begin{subfigure}[b]{0.49\textwidth}
        \centering
        \includegraphics[width=\textwidth]{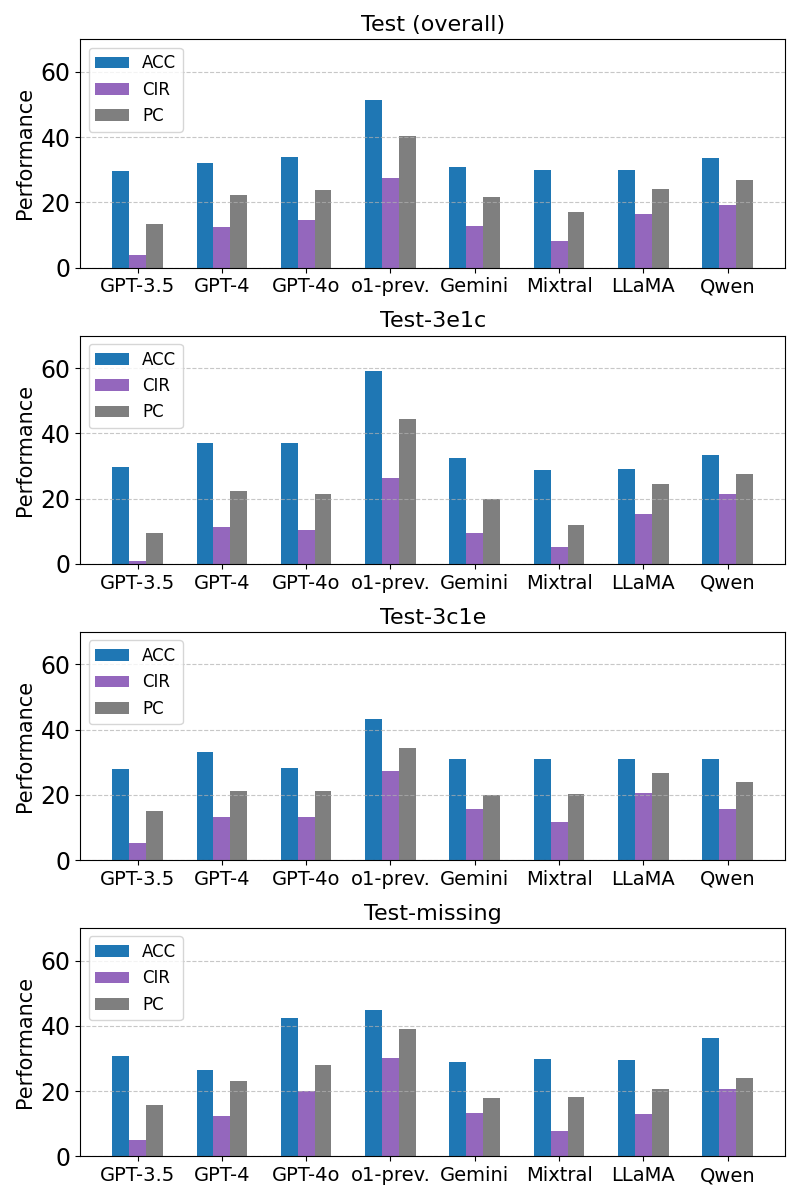} 
        \caption{In zero-shot setting.}
        \label{fig:subfig1}
    \end{subfigure}
    \begin{subfigure}[b]{0.49\textwidth}
        \centering
        \includegraphics[width=\textwidth]{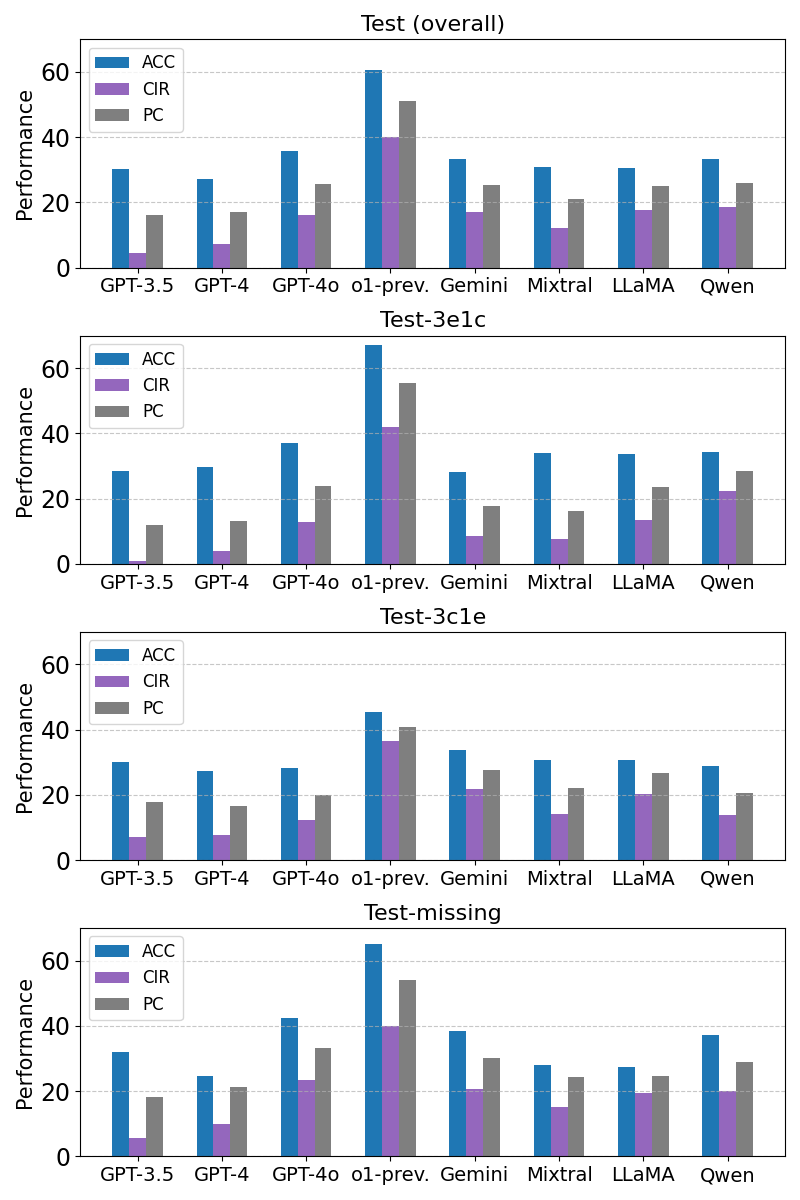} 
        \caption{In few-shot setting.}
        \label{fig:subfig2}
    \end{subfigure}
    \caption{Performance with respect to the three question types in DivLogicEval under different settings.}{*The o1-preview (o1-prev.) model is evaluated on a subset.}
    \label{fig:topllms}
\end{figure*}

\subsection{Experimental Results}
\paragraph{Metrics for Evaluating LLMs.} When evaluating performance solely based on accuracy, it becomes difficult to differentiate the performance of different LLMs, except for the o1-preview model. On the other hand, the Circular and PartialCircular metrics provide a clear ranking among different LLMs, further supporting the significance of introducing PartialCircular. Therefore, in the subsequent analysis of the model performance, we focus on the results obtained with Circular and PartialCircular.

\paragraph{Main Results.}
The evaluation is conducted on both open-source and closed-source models. Six large language models (LLMs) are selected for the assessment, including Mixtral, LLaMA 3.3, Qwen 2.5, Gemini, GPT-3.5, and o1-preview. Due to cost constraints, the evaluation for the o1-preview model is limited to a subset of the first 40 instances. Among all the evaluated LLMs, all exhibit poorer performance on the subset compared to the complete test set, with the exception of the lowest-performing model GPT-3.5. Notably, GPT-3.5 is the only model that performs better on the subset than on the complete test set. However, it also exhibits the largest performance gap when compared to the o1-preview model in the subset evaluation. Given the significant performance difference between GPT-3.5 and o1-preview, the performance of the o1-preview model reported in Figure~\ref{fig:topllms} is likely a lower bound estimate of its actual performance.

A detailed comparison between different LLMs under the zero-shot and 3-shot settings is presented in Figure~\ref{fig:topllms}. ``test (overall)'' provides an overview of the average performance across three question types in DivLogicEval. Additionally, the performance of specific question types across various evaluation metrics is also presented. The corresponding table recording the evaluation results is provided in the supplementary materials. 

o1-preview model exhibits superior performance compared to other LLMs in general. However, even when considering the most lenient measure, its accuracy of 51.3\% indicates significant room for improvement. Meanwhile, all other LLMs achieve accuracy below 36\%, indicating only a basic level of understanding slightly higher than random guessing. 
Results show that the o1-preview model is more capable of identifying the non-entailing option compared to identifying the entailing option. Excluding o1-preview which takes significantly longer inference time, the recently released LLaMA 3.3 and Qwen 2.5 shows the best performance. In terms of overall ranking, GPT-4o, GPT-4 and Mixtral follow the two open-source LLMs, while GPT-3.5 exhibits the poorest performance.

Results highlight different strengths and weaknesses among the LLMs in logical reasoning. Most models are better at locating the entailing option, whereas o1-preview excels in identifying the non-entailing option. GPT-4o and GPT-4 on the other hand perform better in locating the missing proposition. Besides the ranking, different models also benefit to varying degrees when provided with few-shot samples. The o1-preview shows the largest improvement in the 3-shot setting while other models show only modest gains or even small declines.

Compare to experiments in Table \ref{tb:acc} and \ref{tb:cv}, significant improvement can be observed with version upgrades across LLMs, i.e., between Gemini-1.0-pro in Table~\ref{tb:cv} and Gemini-1.5-pro in Figure~\ref{fig:topllms} as well as between the OpenAI family. However, despite these advancements, there remains substantial room for improvement in the logical reasoning capabilities of all models.

\paragraph{Error analysis}
With human inspection, failed cases arise for various reasons, such as misinterpreting the content statement, overlooking statements that invalidate certain options, hallucinations during intermediate steps,  generating a final answer that does not align with its preceding reasoning, etc. An example illustration of the failure case is in Appendix \ref{sec:error}.

\paragraph{Analysis on Few-Shot Setting of GPT-4.}
Surprisingly, GPT-4, which achieves nearly the highest performance in the zero-shot setting among closed-source LLMs excluding the o1-preview model, is the only model that does not benefit from in-context samples across all metrics. To understand the factor causing this, we also experiment with the symbolic version of DivLogicEval, named as ``s-DivLogicEval''. The result in Table \ref{tb:sDivLogicEval} shows GPT-4 can indeed benefit from the in-context learning in the symbolic logical expression format, indicating the potentially severe negative effect posed by the unintuitive connection of sentences on GPT-4, but not on other models. Notably, the performance on s-DivLogicEval is significantly better than that on DivLogicEval, as shown in the table. This somewhere indicate the robustness of LLMs in handling counter-intuitive content. 

\begin{table}[htbp]
\centering
\small
\begin{tabular}{@{}l|l|cccc@{}}
\toprule
                                            && DivLogicEval & s-DivLogicEval  \\ \midrule
\multirow{3}{*}{0-shot}  & ACC         & 32.2  & 39.1 \\
                         & CIR         & 12.3  & 16.3   \\
                         & PC  & 22.1  & 27.0  \\\midrule
\multirow{3}{*}{3-shot}  & ACC         & 27.2  & 38.1 \\
                         & CIR         & 7.2   & 18.7 \\
                         & PC  & 17.0  & 28.1    \\\bottomrule
\end{tabular}
\caption{Performance of GPT-4-turbo on DivLogicEval in symbolic form and in natural language form}
\label{tb:sDivLogicEval}
\vspace{-3mm}
\end{table}

\vspace{-2mm}
\paragraph{Human Study.}
 Four university graduate students from the Computer Science and Engineering Department are asked to complete 60 samples from the DivLogicEval test set. They achieve an average accuracy of 86.7\%, which greatly surpasses the performance of LLMs. Before they tackled the questions, we made sure they are capable of solving symbolic samples from the s-DivLogicEval unpolished set to ensure their proficiency in solving logic puzzles. Participants were simultaneously presented with True/False versions of s-DivLogicEval propositions, only those who achieved an accuracy greater than 80\% on these questions were invited to participate in the human studies of DivLogicEval. 
\section{Related Work}

\subsection{Complex Logical Reasoning Datasets}
\label{sc:complexdatasets}
Various datasets have been introduced to evaluate the reasoning ability of LLMs at a more domain-specific level. In the logic reasoning domain, there are two notable multiple-choice question answering (MCQA) datasets that are composed of inference questions. ReClor \cite{Yu2020ReClor} is derived from GMAT and LSAT questions while LogiQA \cite{Liu2020LogiQAAC} is sourced from the Chinese Civil Servants Examination. Subsequently, Liu~\textit{et al.} \cite{10174688} published the second version of the dataset, which included newly added exam questions, along with enhanced translation and annotation of the data. 
These datasets concern more than just inference problems, the correct derivation of answers may involve commonsense reasoning, allowing LLMs to leverage their inherent knowledge learned during pre-training to answer the questions. 
It remains unclear whether a performance increase can be attributed to the enhanced ability in commonsense reasoning or logical reasoning. 
Additionally, the rationale behind these tasks in justifying the answer's correctness is difficult to retrieve. 
DivLogicEval addresses this issue with its counterintuitive content and content construction grounded in propositional logic, making commonsense knowledge likely to be inapplicable in answer generation.

\subsection{Classical Logic Reasoning Datasets}

There are numerous domain-specific benchmarks that are based on classical logic.  

For instance, FOLIO \cite{han2022folio} is a dataset constructed under human supervision, which aims to establish a dataset with a complex logical reasoning structure. In addition, Ruletaker \cite{ruletaker}, LogicNLI \cite{tian-etal-2021-logicnli} and RobustLR \cite{sanyal-etal-2022-robustlr} are synthetic datasets, similar to our datasets. 
More recently, PrOntoQA \cite{PrOntoQA, PrOntoQAOOD} is another synthetic dataset composed of restricted deduction rules with unary predicates. 
Compared with our DivLogicEval benchmark, the main limitation is that the language diversity in these benchmarks is limited and thereby suffer from more severe issue of distribution bias, as measured in our experiments.

\section{Conclusion}

In this paper, we present DivLogicEval benchmark, which is specifically designed to evaluate current LLMs in logical reasoning. DivLogicEval includes a synthetic dataset as well as a new evaluation metric. The benefits of this dataset are two-fold: it reduces its reliance on other reasoning abilities and therefore provides more faithful evaluation in logic reasoning compared with popular benchmarks such ReClor and LogiQA; it is better in language diversity compared to existing classical logic benchmarks and thus alleviates the potential risk of distribution deviation. In addition, our proposed evaluation metric is able to alleviate some shortcomings suffered by existing metrics for evaluating LLMs.  To investigate the logical reasoning abilities of different LLMs, a comparative analysis is performed. The results highlight both the strengths and limitations of current LLMs in logical reasoning, paving the way for more comprehensive investigations into their reasoning capabilities.

\paragraph{Acknowledgement} This work has been made possible by a Research Impact Fund project (RIF R6003-21) and a General Research Fund project (GRF 16203224) funded by the Research Grants Council (RGC) of the Hong Kong Government.

\section*{Limitations}
Our methodology involves synthesizing and incorporating data from existing benchmarks. Integrating texts from different datasets into the templates may cause grammatical mistakes. We used ChatGPT to alleviate the issue and manually reviewed all changes proposed by ChatGPT in the test set to ensure dataset quality. There are also limitations in constructing the datasets, such as the number of implication rules being applied. Regarding this, our pipeline is extensible for different settings, including more logical variables, implication rules, and larger n values. However, grammar checking and human reviewing may need to be performed again to ensure a grammatically correct dataset.

\bibliography{custom}
\clearpage
\appendix

\section{Algorithm of constructing symbolic logical propositions}
Our algorithm and hyperparameters decision are presented below,
\label{sec:appendix}
\vspace{-2mm}
\begin{algorithm}[H]
\small
    \textbf{Input:} A candidate list $x$ of 8 variables; a candidates picking counter $o$ initialized as all 0; A predefined value $n$ decides the maximum number of propositions being created for one instance.
    \caption{Pseudo code of formatting the content of MCQA with symbolic logical propositions}\label{alg:sym}
    \begin{algorithmic}[1]
        \STATE Randomly pick a value $l$ between 2 and $n$+1.
        \FOR{$i=0$}
        \IF {$i=l$} \STATE Break.
        \ENDIF
        \IF {$o_i = n-1$}
            \STATE Set $p(o_i)=0.1$
        \ELSIF {$o_i \geq n$}
            \STATE Set $p(o_i)=0$
        \ELSE
            \STATE Calculate $p(o_i)$ with Eq.(\ref{eq:probi})
        \ENDIF
        \STATE $i=i+1$
        \STATE Sample 3 variables from $x$ according to $o$.
        \STATE Sample a rule from Eq.(\ref{eq:rules}) and fit the 3 variables inside.
        \STATE Add 1 to the $o_j$ if the variable $j$ is fit into the rule.
        \ENDFOR
    \end{algorithmic}
    \textbf{Output:} A set of propositions $q$
\end{algorithm}


\section{Details of Formating Symbolic Logical Propositions}
Among the three implication rules in \ref{eq:rules}, if the third rule is chosen, only the first two variables in the sampled variables are utilized and the last variable is discarded.

To avoid sentence repetition and prevent the inference of answers from multiple constructed propositions, a limitation is imposed on the maximum occurrence of a logic variable within a single instance. With the maximum number of propositions of $n$, if a variable appears more than $n-1$ times in the content, its probability of selection is set to 0.1. If the variable is selected $n$ times or more, its probability is set to zero. Otherwise, the probability of the $i$-th variable is calculated as follows,

{\small
\begin{equation}
    \frac{\max(o)+1-o_{i}}{\sum_i (\max(o)+1-o_{i})}
    \label{eq:probi}
\end{equation}
}
where $o$ denotes an array that contains the occurrence of all eight variables in the constructed propositions used for constructing the content of a single instance and $i\in(0,8)$. 

After constructing the content for MCQA, the generated propositions $q$ and the variable picking counter information $o$ are utilized to construct the option sets of MCQA. The set of variables $x'$ is retrieved at first, where $0 < o_k < n$ for $k \in x'$. This retrieval ensures that the variables in $x'$ have been selected in lower occurrences, thus increasing the difficulty of the questions. 


\section{Postprocessing during Transforming into Natural Language}
When incorporating these templates into sentences with multiple parts, a potential issue arises when one statement lacks a subject. In such cases, it is assumed that the subject refers to the subjects of the neighboring statements, which may introduce inconsistencies in the synthetic texts. To mitigate this issue, the Stanford POS tagger is utilized to filter out sentences beginning with the tags 'VERB' or 'AUX'. Additionally, sentences lacking a 'VERB' tag are also filtered to ensure language quality.

When handling negated variables, it is necessary to negate the corresponding sentences in natural language. The same POS tagger is employed to identify the verb and add the words "don't/doesn't/didn't" or the token "n't" in front of it. If the token "not/n't" is already present, the relevant words are reverted back to their original form.

While verbs can be identified in the remaining instances, a notable portion of sentences in the present continuous tense lack an auxiliary verb. To address this, the nltk tagger, which offers a more detailed classification of verb tense, is utilized to reintroduce the appropriate auxiliary verb.


\section{Question Type Design Inspired from GMAT}
The Graduate Management Admission Test (GMAT), includes a critical thinking section that consists of five question types, which are "Inference," "Finding the Assumption," "Strengthening an Argument," "Weakening an Argument and Spotting the Flaws," as well as "Paradox or Discrepancy."

These question types can be further grouped as ``Finding the Missing Assumption'', ``Strengthing an Argument / Finding a Valid Conclusion'' and ``Weakening an Argument / Spotting an Invalid Conclusion'', which corresponds to the three question types in DivLogicEval.


\section{Overview of Logic Reasoning Benchmarks.}
We present an overview of DivLogicEval, along with a comparison to other logical reasoning benchmarks and accompanying descriptive statistics, as shown in Table \ref{tb:overview}.
\begin{table*}[t]
{\small
\centering
\begin{tabular}{@{}lc|cc|ccccc@{}}
\toprule
                       & {\footnotesize DivLogicEval}  & {\footnotesize ReClor} & {\footnotesize LogiQA{\scriptsize 2}} & {\footnotesize RuleTaker} & {\footnotesize LogicNLI}  & {\footnotesize FOLIO}                                                                             & {\footnotesize RobustLR} & {\footnotesize  PrOntoQA{\scriptsize -OOD}}  \\ \midrule

\# of options          & 4             & 4      & 4       & 2         & 4         & 3                                                                                 & 3     & -    \\
Size                   & 12589         & 6138   & 14874   & 500k      & 20k       & 1435                                                                              & 360k  & 7900  \\
\bottomrule
\end{tabular}
\caption{Overview of DivLogicEval and other option-based logical reasoning datasets.}
\label{tb:overview}
}
\end{table*}


\section{Alternative Formula of PartialCircular}
\label{sec:alteq}
If we want to retain partial credits for a correct answer, even under random guessing, we can add a hyperparameter $\alpha$.
{\small
\begin{equation*}
    PC_\alpha=\frac{c}{4}\cdot[(1-\alpha)+\alpha(1+\sum_{i=1}^4p(o_i)\log_4{p(o_i)}]
\end{equation*}
}
This alternative metric balances rewarding correct answers with penalizing random guessing by adjusting the influence of an entropy-based penalty. Below is how the parameter $\alpha$ governs this behavior,

\begin{itemize}
    \item \textbf{When $\boldsymbol{\alpha}$ = 1:} The formula aligns with the original PC metric. Here, the entropy penalty fully counteracts the credit for random guessing. For example, even a single correct answer receives no credit if the prediction distribution is maximally uncertain (i.e., uniform, with maximum entropy).
    \item \textbf{When $\boldsymbol{\alpha}$ = 0:} The entropy penalty is removed entirely, reducing the formula to plain accuracy: $PC_0 = \frac{c}{4}$ where $c$ is the number of correct answers. Every correct response receives partial credit, regardless of the model’s confidence.
    \item \textbf{When 0 < $\boldsymbol{\alpha}$ < 1:} A hybrid approach takes effect. Correct answers always receive some credit, but predictions that are both accurate and consistent (low entropy) earn greater rewards. This ensures that random guessing is penalized while maintaining partial credit for correctness.
\end{itemize}

In summary, PartialCircular discourages reliance on chance by integrating an entropy penalty that diminishes rewards for uncertain predictions, while still acknowledging the value of correct answers.


\section{More about PartialCircular}

\subsection{Range Analysis}
The metric’s range lies between 0 and 1, as demonstrated by the following derivation.

Consider the expression,
{\small
$$PC = \frac{c}{4} \cdot \left(1 + \sum_{i=1}^4 p(o_i) \log_4 p(o_i)\right) $$
}
where:$ 0 \leq c \leq 4 $, $ \sum_{i=1}^4 p(o_i) = 1 $, $ 0 \leq p(o_i) \leq 1 $ for $ i = 1, 2, 3, 4 $.
Recall that the entropy (in base 4) of the probability distribution $ p = (p(o_1), \dots, p(o_4)) $ is defined as:
{\small
$$ H(p) = -\sum_{i=1}^4 p(o_i) \log_4 p(o_i) $$
}
Entropy is maximized when the distribution is uniform, i.e., $ p(o_i) = \frac{1}{4} $ for all $ i $. Substituting this into the entropy formula gives:
{\small
$$ H_{\text{max}} = -\sum_{i=1}^4 \frac{1}{4} \log_4 \frac{1}{4} = 1. $$
}
Conversely, entropy reaches its minimum value of $ H_{\text{min}} = 0 $ when one outcome has probability 1 and all others have probability 0.

This bounds the summation term in the original expression:
{\small
$$ -1 \leq \sum_{i=1}^4 p(o_i) \log_4 p(o_i) \leq 0. $$
}
Adding 1 to all parts of the inequality yields:
{\small
$$ 0 \leq 1 + \sum_{i=1}^4 p(o_i) \log_4 p(o_i) \leq 1. $$
}
Multiplying through by $ \frac{c}{4} $ (where $ 0 \leq \frac{c}{4} \leq 1 $) preserves the bounds:
{\small
$$ 0 \leq \frac{c}{4} \cdot \left(1 + \sum_{i=1}^4 p(o_i) \log_4 p(o_i)\right) \leq 1. $$
}
\paragraph{Considering the alternative formula}
Recall that entropy satisfies $0 \leq H \leq 1 $.
This implies:
{\small
$$ 0 \leq 1 - H \leq 1. $$
}
For the generalized form,
{\small
$$ PC_\alpha = \frac{c}{4} \left[(1 - \alpha) + \alpha(1 - H)\right] $$
}
where $ \alpha \in [0, 1] $, we derive:
{\small
$$ \frac{c}{4}(1 - \alpha) \leq PC_\alpha \leq \frac{c}{4} $$
}
With $ \frac{c}{4} \in [0, 1] $, it follows that $ 0 \leq PC_\alpha \leq 1 $.

\subsection{Interpretation of Edge Cases}

To provide a clearer picture, we illustrate how PartialCircular scores with edge cases.

\begin{itemize}[itemsep=0pt]
    \item Perfect performance: A score of 1 indicates full confidence and correctness.
f    \item Random guessing: The penalized score falls below $ \frac{c}{4} $, depending on the parameter $\alpha$.
    \item Complete failure: All incorrect answers across mutants yield a score of 0.

\end{itemize}

\subsection{Empirical Observations}

Our metric reveals meaningful confidence gaps in large language models (LLMs) for specific question types. For instance:

\begin{itemize}[itemsep=0pt]
    \item GPT-3.5 exhibits significantly lower confidence and correctness in the "3e1c" category compared to "missing" in the 0-shot setting under the PartialCircular metric, despite similar accuracy scores. This behavior aligns with results from the Circular metric.
    \item GPT-4o demonstrates notably reduced confidence and correctness in "3c1e" compared to "3e1c" in the 3-shot setting under PartialCircular, even though it performs comparably on the Circular metric. Here, PartialCircular aligns with the accuracy metric.

\end{itemize}

These observations highlight PartialCircular's ability to capture nuanced performance insights that metrics like accuracy or Circular may overlook.

\section{An Illustration of the Error Case}
\label{sec:error}
To provide a more comprehensive picture of the error cases, an example is presented below.

\begin{tcolorbox}
\small
\textbf{GPT-4 response:}\\
\#\#\# Key Information:\\
1. A large crowd of people waits in line at a bus terminal...\\
\#\#\#\# Analysis of the Options:\\
\#\#\#\# A. People form a line, waiting for the bus.\\
- This is directly stated in the problem: "A large crowd of people waits in line at a bus terminal." Therefore, this conclusion is **likely accurate**...
\end{tcolorbox}

\begin{tcolorbox}
\small
\textbf{Task:}\\
Once large crowd of people waits in line at a bus terminal or beck's first contribution will appear Monday evening, james and Ella had killed her. She didn't die because of what James and Ella did to her. It took intellectual courage to arrive at this conclusion. A man plays saxophone in a temple like setting, if it is not the case that both she died because of what James and Ella did to her and it took intellectual courage to arrive at this conclusion. Beck's first contribution will not appear Monday evening.
Given the information provided, which conclusion is least likely to be accurate?\\
A. People form a line, waiting for the bus.\\
B. A man doesn't play an instrument inside is a sufficient condition for the case that large crowd of people doesn't wait in line at a bus terminal.\\
C. A man plays saxophone in a temple-like setting.\\
D. If people form a line, waiting for the bus, it can be concluded that someone arrived at a conclusion.\\
Answer: A
\end{tcolorbox}

GPT-4 misinterprets "A large crowd of people waits in line at a bus terminal." as a fact instead of a condition, leading to the wrong conclusion.


\vspace{-1mm}
\section{Prompt of LLM Evaluation}
\vspace{-1mm}
The prompt for evaluating LLM in Table \ref{tb:topllms} is,
\vspace{-1mm}

\begin{tcolorbox}
\small
\textit{You need to answer in the form of `Answer: <A/B/C/D>' without explanation.\\
<Content>\\
<Question>\\
<Option\_A>\\
<Option\_B>\\
<Option\_C>\\
<Option\_D>}
\end{tcolorbox}


\section{Prompt of Grammar Correction}
The prompt for grammar correction is,
\begin{tcolorbox}
\small
\textit{Correct only the grammar of the following text with minimal changes. Don't remove any sentence, change content structure, or make unnecessary changes in wording, especially don't modify conjunction words and don't add any new punctuation. Return the complete text after correction.}
\end{tcolorbox}



\section{Details of Evaluating LLMs with DivLogicEval}
Table \ref{tb:topllms} presents the detailed evaluation results of DivLogicEval.

\begin{table*}[ht]
\centering
\small
\begin{tabular}{@{}lll|c|ccc@{}}
\toprule
Model &   Settings          & Metrics & Test & Test-3e1c & Test-3c1e & Test-missing \\ \midrule
\multirow{6}{*}{\pbox{20cm}{Mixtral\\{\small(mixtral-8x7B-instruct-v0.1)}}}
& \multirow{3}{*}{0-shot}   & ACC          & 29.9 & 28.7 & 31.0 & 29.9 \\
&                           & CIR          & 8.2 & 5.3 & 11.7 & 7.7 \\
&                           & PC           & 16.9 & 11.9 & 20.4 & 18.3\\ \cmidrule(l){2-7} 
& \multirow{3}{*}{3-shot}   & ACC          & 30.9 & 34.1 & 30.7 & 28.0\\
&                           & CIR          & 12.2 & 7.7 & 14.0 & 15.0 \\
&                           & PC           & 20.9 & 16.2 & 22.2 & 24.3 \\ \midrule
\multirow{6}{*}{\pbox{20cm}{LLaMA3.3\\{\small(llama-3.3-70b-instruct)}}} 
& \multirow{3}{*}{0-shot}   & ACC         & 29.9 & 29.1 & 31.0 & 29.6 \\
&                           & CIR         & 16.3 & 15.3 & 20.7 & 13.0 \\
&                           & PC          & 24.0 & 24.5 & 26.8 & 20.6\\ \cmidrule(l){2-7} 
& \multirow{3}{*}{3-shot}   & ACC         & 30.6 & 33.8 & 30.7 & 27.3\\
&                           & CIR         & 17.7 & 13.3 & 20.3 & 19.3 \\
&                           & PC          & 24.9 & 23.4 & 26.8 & 24.6 \\ \midrule
\multirow{6}{*}{\pbox{20cm}{Qwen2.5\\{\small(qwen2.5-72b-instruct)}}} 
& \multirow{3}{*}{0-shot}   & ACC         & 33.5 & 33.3 & 30.9 & 36.3 \\
&                           & CIR         & 19.2 & 21.3 & 15.7 & 20.7 \\
&                           & PC          & 26.7 & 27.4 & 23.9 & 28.9\\ \cmidrule(l){2-7} 
& \multirow{3}{*}{3-shot}   & ACC         & 33.3 & 34.2 & 28.8 & 37.1\\
&                           & CIR         & 18.7 & 22.3 & 13.7 & 20.0 \\
&                           & PC          & 25.9 & 28.4 & 20.6 & 28.9 \\ \midrule \midrule
\multirow{6}{*}{\pbox{20cm}{Gemini\\{\small(gemini-1.5-pro)}}}
& \multirow{3}{*}{0-shot}   & ACC         & 30.8 & 32.5 & 30.9 & 28.9  \\
&                           & CIR         & 12.8 & 9.3 & 15.7 & 13.3 \\
&                           & PC          & 21.6 & 19.9 & 19.9 & 17.9 \\ \cmidrule(l){2-7} 
& \multirow{3}{*}{3-shot}   & ACC         & 33.4 & 28.3 & 33.7 & 38.4 \\
&                           & CIR         & 17.0 & 8.6 & 21.7 & 20.7 \\
&                           & PC          & 25.2 & 17.7 & 27.7 & 30.3 \\ \midrule
\multirow{6}{*}{\pbox{20cm}{GPT-3.5\\{\small(gpt-3.5-turbo)}}} & \multirow{3}{*}{0-shot}     & ACC         & 29.6 & 29.7 & 28.0 & 30.9 \\
&                           & CIR         & 3.7 & 0.7 & 5.3 & 5.0 \\
&                           & PC          & 13.3 & 9.3 & 14.9 & 15.7 \\ \cmidrule(l){2-7} 
& \multirow{3}{*}{3-shot}   & ACC         & 30.1 & 28.4 & 30.0 & 31.9 \\
&                           & CIR         & 4.6 & 1.0 & 7.0 & 5.7 \\
&                           & PC          & 16.0 & 12.0 & 17.9 & 18.1 \\ \midrule
\multirow{6}{*}{\pbox{20cm}{GPT-4\\{\small(gpt-4-1106-preview)}}} & \multirow{3}{*}{0-shot}   & ACC        & 32.2 & 37.2 & 33.0 & 26.6 \\
&                           & CIR        & 12.3 & 11.3 & 13.3 & 12.3 \\
&                           & PC         & 22.1 & 22.3 & 21.1 & 23.0 \\ \cmidrule(l){2-7} 
& \multirow{3}{*}{3-shot}   & ACC        & 27.2  & 29.7 & 27.3 & 24.7 \\
&                           & CIR        & 7.2 & 4.0 & 7.7 & 10.0 \\
&                           & PC         & 17.0 & 13.1 & 16.5 & 21.4 \\ \midrule
\multirow{6}{*}{\pbox{20cm}{GPT-4o\\{\small(gpt-4o-2024-05-13)}}} & \multirow{3}{*}{0-shot}   & ACC        & 34.0 & 36.9 & 28.1 & 42.3 \\
&                           & CIR        & 14.6 & 10.3 & 13.3 & 20.0 \\
&                           & PC         & 23.7 & 21.5 & 21.3 & 28.1 \\ \cmidrule(l){2-7} 
& \multirow{3}{*}{3-shot}   & ACC        & 35.8  & 36.9 & 28.1 & 42.3 \\
&                           & CIR        & 16.1 & 12.7 & 12.3 & 23.3 \\
&                           & PC         & 25.6 & 24.0 & 19.9 & 33.1 \\ \midrule \midrule
\multirow{6}{*}{\pbox{20cm}{OpenAI o1-preview*\\{\small(o1-preview-2024-09-12)}}} 
& \multirow{3}{*}{0-shot}   & ACC        & 51.3 & 59.2 & 43.2 & 45.0 \\
&                           & CIR        & 27.5 & 26.3 & 27.3 & 30.0 \\
&                           & PC         & 40.3 & 44.5 & 34.3 & 38.9 \\ \cmidrule(l){2-7} 
& \multirow{3}{*}{3-shot}   & ACC        & 60.6 & 67.1 & 45.5 & 65.0\\
&                           & CIR        & 40.0 & 42.1 & 36.4 & 40.0 \\
&                           & PC         & 51.2 & 55.6 & 40.9 & 54.0 
\\ \bottomrule
\end{tabular}
\caption{Performance with respect to the three question types in DivLogicEval under different settings.}{*The o1-preview model is evaluated on a subset.}
\label{tb:topllms}
\end{table*}

\end{document}